\documentclass[10pt,twocolumn,letterpaper]{article}

\usepackage[pagenumbers]{cvpr} 

\newcommand{\red}[1]{{\color{red}#1}}

\definecolor{cvprblue}{rgb}{0.21,0.49,0.74}
\usepackage[pagebackref,breaklinks,colorlinks,allcolors=cvprblue]{hyperref}
\usepackage{outline}
\usepackage[utf8]{inputenc} 
\usepackage[T1]{fontenc}    
\usepackage{hyperref}       
\usepackage{url}            
\usepackage{booktabs}       
\usepackage{amsfonts}       
\usepackage{nicefrac}       
\usepackage{pmgraph}
\usepackage[normalem]{ulem}
\usepackage[utf8]{inputenc}
\usepackage{amsbsy,amsthm,amsmath,amssymb,pict2e,picture} 
\hypersetup{colorlinks,allcolors=black}
\usepackage{adjustbox}

\usepackage{graphicx}
\usepackage{times}
\usepackage{xcolor}
\usepackage{xspace}
\usepackage{bm} 

\usepackage{caption}
\usepackage{subcaption}
\usepackage{epstopdf}

\usepackage{enumitem}
\usepackage{graphicx}
\usepackage{epsfig}
\usepackage{tikz}
\usepackage{algpseudocode}
\usepackage{algorithm}
\usepackage{mathrsfs}
\usepackage{setspace}

\long\def\comment#1{}

\newcommand{\der} {\xmath{\mathrm{d}}}

\newcommand{\xmath}[1] {\ensuremath{#1}\xspace}
\newcommand{\blmath}[1] {\xmath{\bm{#1}}}
\newcommand{\A}{\blmath{A}}

\newcommand{\D}{\blmath{D}}

\newcommand{\G}{\blmath{G}}
\newcommand{\I}{\blmath{I}}

\newcommand{\w}{\blmath{w}}
\newcommand{\x}{\blmath{x}}

\newcommand{\s}{\blmath{s}}
\newcommand{\y}{\blmath{y}}
\newcommand{\n}{\blmath{n}}

\renewcommand{\u} {\blmath{u}}
\renewcommand{\v} {\blmath{v}}

\newcommand{\bu}{\blmath{u}}
\newcommand{\bv}{\blmath{v}}

\newcommand{\bI}{\boldsymbol{I}}

\newcommand{\bepsilon}{\blmath{\epsilon}}

\long\def\red#1{\bgroup\color{red}#1\egroup}
\long\def\purple#1{\bgroup\color{purple}#1\egroup}

\definecolor{mich-blue-high}{HTML}{0027CC}
\long\def\blue#1{\bgroup\color{mich-blue-high}#1\egroup}
\long\def\red#1{\bgroup\color{red}#1\egroup}

\makeatletter
\newcommand{\pictslash}[2]{%
  \vcenter{%
    \sbox0{$\m@th#1\varobslash$}\dimen0=.55\wd0
    \hbox to\wd 0{%
      \hfil\pictslash@aux#2\hfil
    }%
  }%
}
\newcommand{\pictslash@aux}[2]{%
    \begin{picture}(\dimen0,\dimen0)
    \roundcap
    \put(0,#1\dimen0){\line(1,#2){\dimen0}}
    \end{picture}%
}
\makeatother

\makeatletter
\newcommand*{\rom}[1]{\expandafter\@slowromancap\romannumeral #1@}
\makeatother

\def\paperID{} 
\def\confName{CVPR}
\def\confYear{2026}

\title{Local Patches Meet Global Context: Scalable 3D Diffusion Priors
\\
for Computed Tomography Reconstruction}

\author{Taewon Yang\thanks{Equal contribution}\quad 
Jason Hu\footnotemark[1]\quad 
Jeffrey A. Fessler\quad 
Liyue Shen\\
EECS Department,
University of Michigan\\
Ann Arbor, MI, 48109\\
{\tt\small \{taewony, jashu, fessler, liyues\}@umich.edu}
}

\begin{document}
\maketitle

\pagestyle{plain}
\begin{abstract}
Diffusion models learn strong image priors
that can be leveraged to solve inverse problems like medical image reconstruction. 
However, for real-world applications such as 3D Computed Tomography (CT) imaging, directly training diffusion models on 3D data presents significant challenges due to the high computational 
demands of extensive GPU resources and large-scale datasets. 
Existing works mostly reuse 2D diffusion priors to address 3D inverse problems,
but fail to fully realize and leverage the generative capacity of diffusion models for high-dimensional data.
In this study, we propose a novel 3D patch-based diffusion model that can learn a fully 3D diffusion prior from limited data, enabling scalable generation of high-resolution 3D images.
Our core idea is to learn the prior of 3D patches to achieve scalable efficiency, while coupling local and global information to guarantee high-quality 3D image generation, by modeling the joint distribution of position-aware 3D local patches and downsampled 3D volume
as global context.
Our approach not only enables high-quality 3D generation, but also 
offers an unprecedentedly efficient and accurate solution to high-resolution 3D inverse problems.
Experiments on 3D CT reconstruction across multiple datasets show that our method outperforms state-of-the-art methods in both performance and efficiency, notably achieving high-resolution 3D reconstruction of $512 \times 512 \times 256$ ($\sim$20 mins).
\end{abstract}    
\section{Introduction}
\label{sec:intro}
\begin{figure}[t]
\centering
\includegraphics[width=1\linewidth]{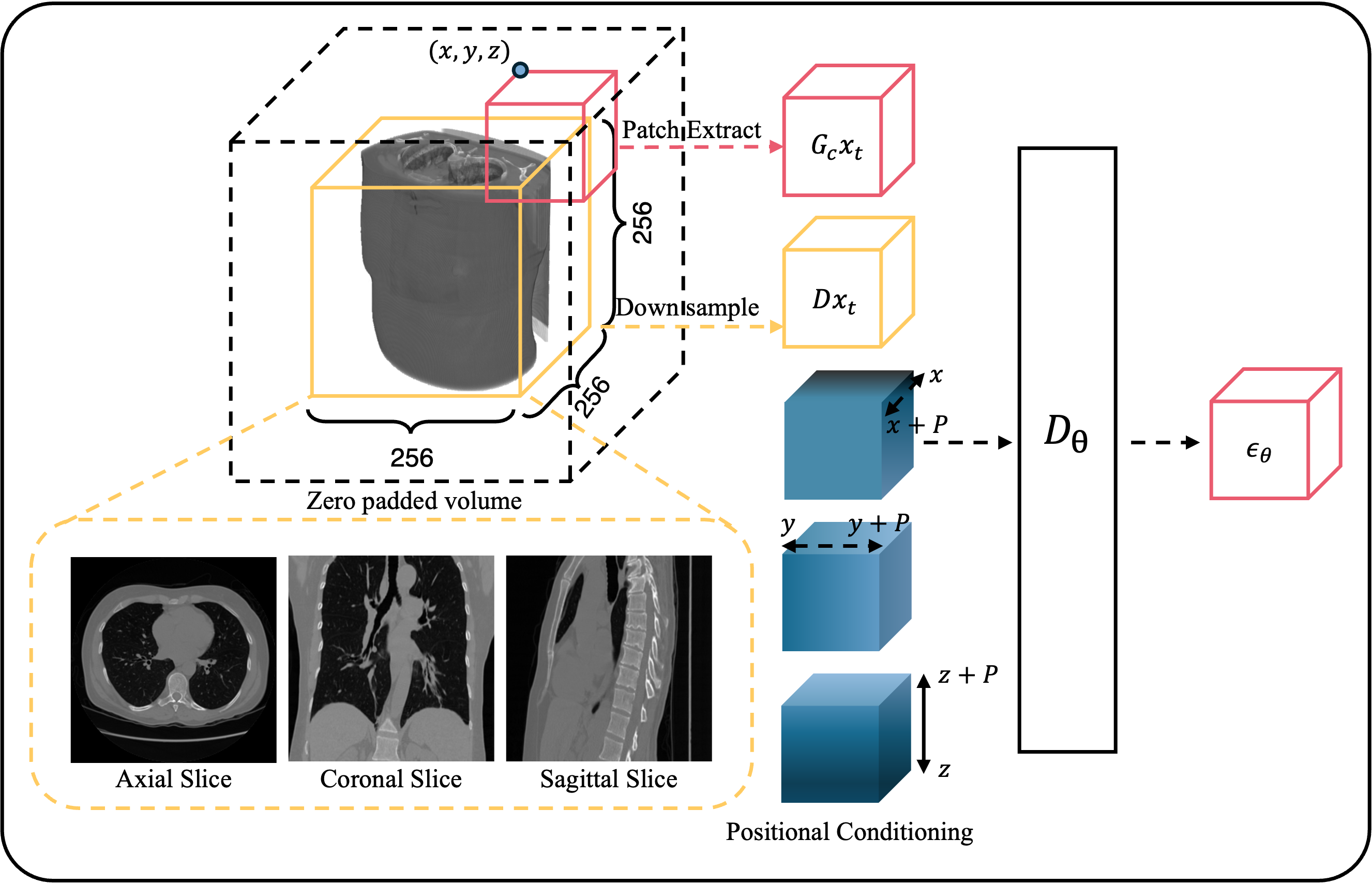}
\caption{Training the 3D patch diffusion model.
Noisy 3D patch $\G_c\x_t$ (local details),
downsampled volume $\D \x_t$ (global context),
and positional encoding are concatenated as inputs
to the denoiser network $D_{\theta}$ in each training iteration,
to predict the noise $\bepsilon_\theta$ in the input noisy 3D patch.
Positional encoding consists of voxel-based coordinates in $x,y,z$
that are normalized to $[-1,1]$.}
\label{fig:Training}
\end{figure}

Diffusion models have emerged as a powerful class of generative model for image generation
\cite{song:21:sbg,improved_ddpm, diffusiontransformer, rombach2022highresolution},
modeling the underlying image distribution $p(\x)$
through learned denoising processes across progressively varied noise levels.
The trained network approximates the score function of image distribution
$\s(\x) = \nabla \log p(\x)$.
At inference time, starting from the initialization of Gaussian noise image,
the noise signal is iteratively removed using the trained denoiser network,
ultimately resulting in clean image samples
drawn from the learned distribution prior $p(\x)$. 

Inverse problems can be formulated as $\y = \mathcal{A}(\x) + \n$,
where \y is a known measurement,
$\mathcal{A}$ is the forward operator,
\x is the unknown image to be recovered, and $\n$ is random noise. 
Specifically, computed tomography (CT) reconstruction can also be formulated as an inverse problem
that aims to reconstruct a 3D volume \x from the measurement \y, 
containing projection views of \x acquired from different angles~\cite{ctold1}. 
Furthermore, to minimize patient radiation exposure,
it is desirable to reduce the number of acquired projection views, typically on the order of 100-1000 views, to substantially fewer
\cite{song2024diffusionblendlearning3dimage, chung:22:s3i},
which is called \textit{sparse-view CT reconstruction}.
In this case, the inverse problem becomes ill-posed,
as there are many potential images \x that could correspond to a measurement \y. 
Hence, it is necessary to impose prior knowledge on \x
to obtain a reasonable reconstruction, such as%
~\cite{KONOVALOV2024104491,lahiri2023sparse} in previous works.

Given that diffusion models can learn powerful generative priors over image data, it is natural to leverage these priors to address
sparse-view CT reconstruction problem. 
Previous works~\cite{chung:23:dps, DDRM, DDNM, myown1, chung2022comecloserdiffusefaster}
have proposed various approaches leveraging diffusion models to solve different inverse problems, including 2D CT reconstruction. 
These approaches generally consist of, at inference time,
interweaving the original diffusion model steps with data fidelity enforcement steps
that push the image to be consistent with the measurements. 

However, challenges emerge when directly applying these existing approaches to real-world sparse-view CT task, where the images are inherently three-dimensional volumes.
A clinically standard CT image
have the full size of $512 \times 512 \times 512$,
making it almost infeasible to fit the entire image into the network training due to memory constraints. 
Furthermore, training diffusion models typically requires large-scale datasets~\cite{huijie},
often comprising hundreds of thousands of images
such as ImageNet~\cite{imagenet} or LSUN~\cite{lsun}.
For CT reconstruction, there is a lack of high-quality volumetric CT data in such scale. 
Finally, 
training time of diffusion models on 3D images would be prohibitively expensive.
Thus, the high computational costs in memory, time, and data pose significant challenges to directly train diffusion models on 3D CT images.

To address these challenges, prior works have proposed leveraging 2D image priors, either by combining them with hand-crafted external regularization terms~\cite{chung:22:s3i} or by training an additional 2D image prior along a perpendicular plane~\cite{lee2023improving}. 
However, these approaches tend to produce blurry artifacts because of sensitivity to the hand-crafted regularization term, or require long inference times due to the use of two perpendicular 2D CT priors.
Recently, \citet{song2024diffusionblendlearning3dimage} introduced a new solution to learn the joint distribution
over a small set of 2D slices simultaneously, but the underlying network remains a 2D architecture with multiple channels, requesting significant reconstruction time comparable to that of \citet{chung2024decomposeddiffusionsampleraccelerating}.
However, these methods do not fully realize and exploit the generative capacity of diffusion models for high-resolution 3D volumetric images.

To overcome these limitations, we propose a novel approach that learns scalable 3D diffusion priors by representing 3D volumes as \textbf{position-condition local patches}. 
To ensure high-quality 3D image generation, our study emphasizes the importance of incorporating global context.
Specifically, by jointly modeling the distribution of local patches alongside a downsampled version of the entire volume, our method effectively unifies global contextual information with local image statistics, to obtain improved \textbf{global-aware patch priors}. 
Beyond image generation, the resulting fully 3D prior demonstrates superior performance in large-scale inverse problem solving, such as high-resolution 3D CT reconstruction.
Overall, our contributions are three-fold. 
\begin{itemize}
\item
We propose a novel 3D patch-based diffusion model that learns a fully 3D diffusion prior, enabling the scalable generation of high-resolution 3D images, and offering an unprecedentedly efficient and accurate solution to address high-resolution 3D inverse problems.
    
\item
Our key technical contribution is a framework that models the joint distribution of position-aware 3D local patches and a downsampled 3D volume, enabling scalable learning of a global-contextual 3D patch-based prior by effectively integrating local and global information to achieve high-quality 3D image generation.
    
\item
Extensive experiments on different datasets and different-sized images
show that our proposed method achieves SOTA performance for 3D sparse-view CT reconstruction
in various settings.
\end{itemize}

\section{Background and Related Works}
\label{sec:formatting}

\subsection{Diffusion Models}
Diffusion models define a forward stochastic differential equation (SDE)
that progressively perturbs a clean sample with Gaussian noise~\cite{song:21:sbg}.
For $t \in [0, T]$, let $\x(t) \in \mathbb{R}^d$ evolve as
\begin{equation}
\der \x = -\frac{\beta(t)}{2} \x \, \der t + \sqrt{\beta(t)} \, \der \w,
\end{equation}
where $\beta(t)$ specifies the variance schedule.
Under this process, $\x(0)$ follows the data distribution,
while $\x(T)$ approaches a standard Gaussian.
To synthesize new samples, one solves the corresponding reverse-time SDE~\cite{ANDERSON1982313}:
\begin{equation}
\der \x = \left( -\frac{\beta(t)}{2} - \beta(t) \nabla_{\x_t} \log p_t(\x_t) \right) \der t
+ \sqrt{\beta(t)} \, \der \overline{\w},
\end{equation}
where $\nabla_{\x_t} \log p_t(\x_t)$
is the score function of the noisy data distribution at time $t$.
By training a neural network to approximate this score function,
one can iteratively integrate the reverse SDE starting from Gaussian noise
to produce realistic samples from the learned data prior.

\subsection{Denoising Diffusion Implicit Models (DDIM)}
Diffusion models are inevitably slow,
as they typically require hundreds or thousands of steps to sample one image.
Prior works have explored accelerating the diffusion sampling stage%
~\cite{DDIM, karras2022elucidating, lu2022dpmsolver},
For example, to expedite the sampling process, DDIM~\cite{DDIM} proposed the following update rule
\begin{equation}\label{eq:DDIM_eq}
    \x_{t-1} = \sqrt{\bar{\alpha}_{t-1} } \, \hat{\x}_t
    +\sqrt{1-\bar{\alpha}_{t-1} -\sigma_t^2} \, \hat{\epsilon}_t+\sigma_t\bepsilon_t
\end{equation}
where $\bepsilon_t\sim \mathcal{N}(0,I)$, $\bar\alpha_{t-1}$ is the noise schedule at timestep $t-1$,
$\hat{\bepsilon_t}$ is the predicted noise
and $\hat{\x}_t$ is the denoised estimate
\begin{equation}\label{eq:denoisedxt}
    \hat{\x}_t = \frac{1}{\sqrt{\bar{\alpha}_t}}(\x_t-\sqrt{1-\bar{\alpha}_t} \, \hat{\bepsilon}_t).
\end{equation}
In \cite{DDIM},
the stochasticity is controlled by $\sigma_t$
that is defined as
\begin{equation}\label{eq:sigma}
    \sigma_t = \eta\sqrt{(1-\bar{\alpha}_{t-1}) /
    (1-\bar{\alpha}_{t})}\sqrt{1-\bar{\alpha}_{t}/\bar{\alpha}_{t-1}}
\end{equation}
where $\eta \in [0,1]$ so that $\eta=0$ leads to fully deterministic sampling. 
Alternatively,
\cite{chung2024decomposeddiffusionsampleraccelerating} considers 
\begin{equation} \label{eq:DDS_sigma}
    \sigma_t=\eta\sqrt{1-\bar{\alpha}_{t-1}},
\end{equation}
for which \eqref{eq:DDIM_eq} becomes
\begin{equation}\label{eq:DDS_DDIM_eq}
    \x_{t-1} = \sqrt{\bar{\alpha}_{t-1} }\hat{\x}_t
    +\sqrt{1-\bar{\alpha}_{t-1}}(\sqrt{1-\eta^2}\cdot\hat{\bepsilon_t}+\eta\bepsilon_t).
\end{equation}
Assuming
the estimated $\hat{\bepsilon_t}$ and the stochastic
$\bepsilon\sim \mathcal{N}(0,\I)$ are independent Gaussians,
$\tilde{\bepsilon}=\sqrt{1-\eta^2}\cdot\hat{\bepsilon_t}+\eta \bepsilon_t$ is also a Gaussian.
Hence, $\tilde{\bepsilon}$ can equivalently be represented as a sample from
$\tilde{\bepsilon}\sim \mathcal{N}(0,\I)$.
This preservation of noise variance
ensures that the model samples $\x_{t-1}$ from the marginal distribution $q(\x_{t-1}|\x_0)$.

\subsection{Patch-based Diffusion Models}
To alleviate the computational demands of diffusion model training and inference,
prior works have explored various potential solutions to enhance the model efficiency.
For example, instead of operating directly in the image space,
latent diffusion models~\cite{rombach2022highresolution} perform the diffusion process
in a compressed latent representation space in lower dimension.
While this approach greatly improves efficiency,
it depends on a pretrained encoder–decoder pair~\cite{esser2021taming},
which assumes access to abundant training data and may require retraining when data domains change.
In addition, patch-based diffusion models~\cite{wang2023patch, ding2023patched}
are proposed to train the diffusion model on only smaller image subregions rather than full images,
enabling memory-efficient training for generation tasks.
However, supervised patch diffusion approaches~\cite{xia:23:svb, visionpatch} 
are designed for specific applications
such as adverse weather image restoration and breast CT reconstruction, thus
do not yield a general-purpose unconditional image prior.
Other patch-based formulations~\cite{bieder:24:ddm, luhman:22:idm, diffusiontransformer}
learn such an unconditional prior but still require the full image during inference,
limiting scalability. 
These unconditional patch-based diffusion models are employed as priors
for solving inverse problems. For example, during each sampling steps,
\citet{hu2024learningimagepriorspatchbased} randomly select the starting location of a grid
and then partition the image into patches along this grid. 
In this framework, the full-image prior is approximated as a product of patch distributions,
and data-consistency terms defined by the forward model are incorporated
when updating the sample to solve the inverse problem.

\subsection{3D CT Reconstruction}
Computed tomography (CT) is a widely used medical imaging technique
that captures the internal structure of a 3D object
by transmitting X-rays through it from multiple angles~\cite{ctold1}.
The measurements, denoted \y,
consist of a collection of 2D projection views
obtained by rotating the X-ray source and detector around the object.
Let $\mathcal{A}$ represent the linear forward operator that models the CT acquisition process,
and let \x denote the unknown 3D image to be reconstructed.
The goal of CT reconstruction is to estimate \x from the observed projections \y.
Classical reconstruction techniques include regularization-based formulations
that impose prior assumptions on \x,
as well as likelihood-based optimization methods~\cite{ctold1, ctold2, ctold3, ctold4}.
More recently, deep learning–based approaches have been introduced%
~\cite{fbpconvnet, lahiri2023sparse, sonogashira2020high, whang2023seidelnet},
in which convolutional neural networks such as U-Net~\cite{fbpconvnet} are trained
to map partial-view filtered backprojection (FBP) images to ground truth full-view reconstructions.
Although these approaches can improve reconstruction quality,
they often yield overly smooth outputs
and fail to generalize to out-of-distribution data~\cite{instability}.

\subsection{Diffusion Models for 3D CT Reconstruction}
Diffusion models provide a powerful class of generative priors
capable of synthesizing realistic images directly from noise.
When applied to inverse problems,
diffusion-based approaches typically formulate reconstruction as a posterior sampling process%
~\cite{MCGchung, chung:23:dps, mcgdiff, DDNM, DDRM},
wherein a network learns an unconditional image prior for \x
that can be reused across tasks without retraining.
While such methods have demonstrated strong performance on 2D inverse problems,
extending them to 3D settings remains challenging due to prohibitive memory and data requirements.

For 3D CT reconstruction,
DiffusionMBIR~\cite{chung:22:s3i} and DDS~\cite{chung2024decomposeddiffusionsampleraccelerating}
address the problem by training diffusion models on 2D axial slices.
During reconstruction,
they employ a total-variation (TV) regularizer with posterior sampling
to encourage inter-slice smoothness.
Although TV regularization provides some degree of slice consistency,
it is not data-driven and fails to capture a true 3D prior.
TPDM~\cite{lee2023improving} mitigates this limitation
by learning an additional prior on coronal slices via conditional sampling,
introducing a data-driven way to maintain slice coherence during reconstruction.
However, this approach requires all training volumes to share the same cubic shape
and doubles the computational cost during training time.
In practice, 3D CT scans have varying numbers of slices.
More recently, DiffusionBlend~\cite{song2024diffusionblendlearning3dimage}
leverages pretrained 2D diffusion models
to learn joint statistics across groups of a few adjacent slices,
yet it still does not capture a fully 3D volumetric prior
capable of generating realistic 3D structures.
To address these limitations,
we propose a new approach for learning
a generative 3D image prior
that can be efficiently applied to large-scale CT reconstruction tasks
within practical computational constraints.

\section{Methods}
\subsection{Global-Aware Patch Prior}
\begin{figure}
  \centering
  \includegraphics[width=0.5\textwidth]{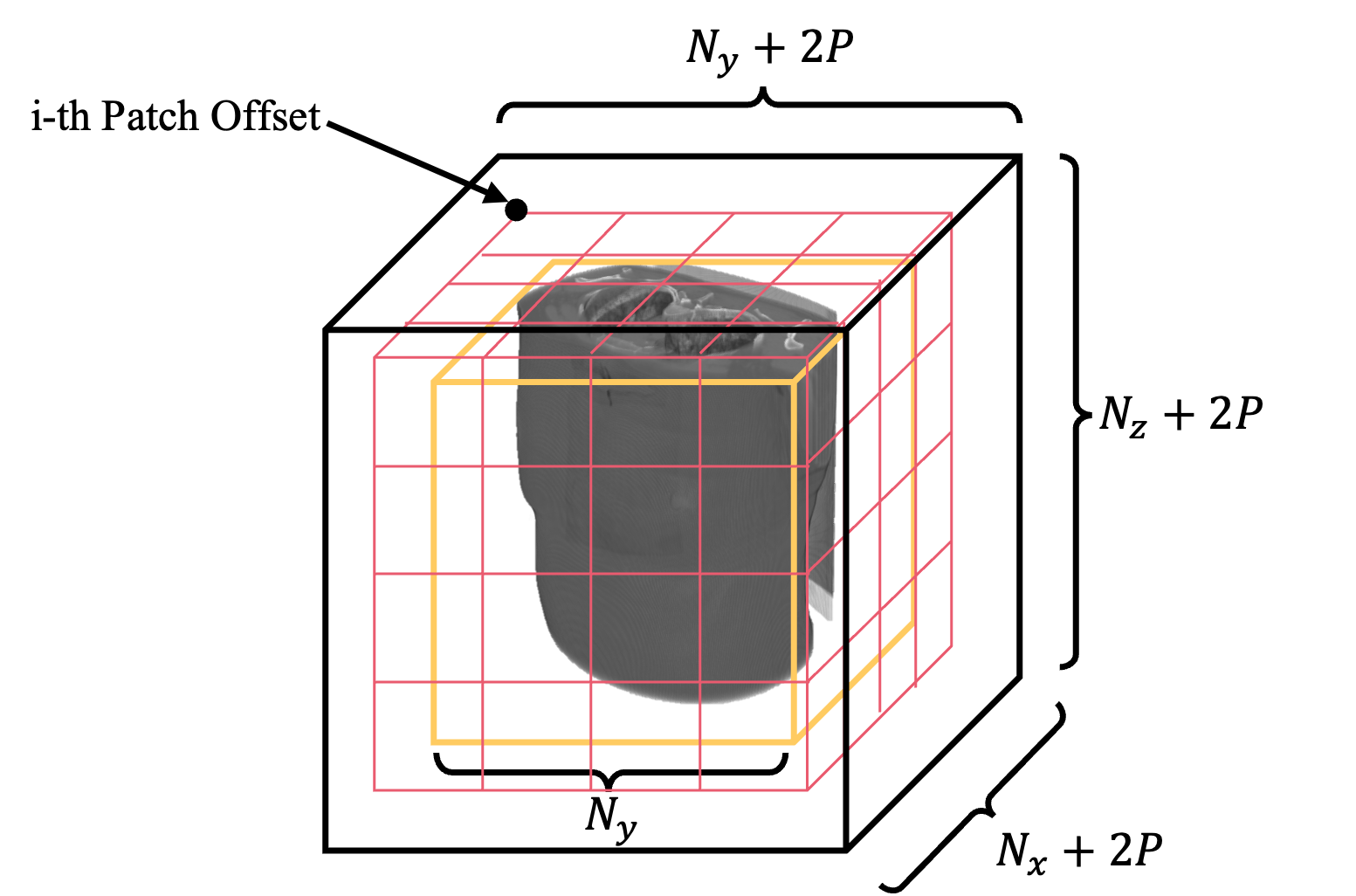}
  \caption{Schematic illustration for zero padding and partitioning image into 3D patches.
  Each index $i$ represents one of $P^3$ possible ways to choose a patch offset tuple.}
  \label{fig:patch_partition}
  \vspace{-18pt}
\end{figure}

We begin by following the formulation of learning patch-based priors as described in~\cite{hu2024learningimagepriorspatchbased}, but generalized for 3D images. 
In patch-based diffusion models, 
one needs first to zero pad
each $N_x \times N_y \times N_z$ image
by an amount $P$ on each side.
Slightly recycling notation,
we let \x denote the resulting padded image. 
We partition \x into many square patches, with one bordering region of zeros by first choosing 
the $i$th patch offset tuple $(o_1,o_2, o_3) \in \{0,\ldots,P-1\}^3$
according to Figure~\ref{fig:patch_partition}.
The number of patches needed in the $x$ direction to perfectly tile the image is  $k_1=N_x/P$, and define $k_2=N_y/P$ and $k_3=N_z/P$ as the number of patches to perfectly tile the image in the $y$ and $z$ directions, respectively. 
Thus, the total number of patches needed to cover the image with an offset is $l = (k_1+1)(k_2+1)(k_3+1)$. 
Hence, our model for the data distribution has the form 
\begin{equation} \label{eq:padis_distribution}
    p(\x) = \frac{1}{Z} \left( \prod_{i=1}^{P^3} p_{i, B}(\x_{i,B}) \prod_{r=1}^{l}
    p_{i,r}(\x_{i,r}) \right)^{1/P^3},
\end{equation}
where $\x_{i,B}$ represents the bordering region of \x
that depends on the specific value of $i$,
$p_{i, B}$ is the probability distribution of that region,
$\x_{i,r}$ is the $r$th $P \times P \times P$ patch
when using the partitioning scheme corresponding to the $i$th patch offset,
$p_{i,r}$ is the probability distribution of that region,
and $Z$ is a normalizing factor.
This model takes a product over all possible patch offsets,
eliminating boundary artifacts that 
would occur if a fixed patch offset was used.

Crucially, however,
the model
\eqref{eq:padis_distribution}
does not account for global structure within the whole image or correlation between two distant patches. 
This can lead to reasonable local structure within each generated local patches,
but a lack of coherent and realistic longer range details. 
At the same time,
memory constraints in 3D preclude
inputting the entire image into the network that approximates any part of the score function. 
Motivated by these considerations, we develop an improved \textbf{patch-based prior with global context} as follows: 
\begin{equation} \label{joint_down}
    p(\x) = \left( \prod_{i=1}^{P^3} p_{i, B}(\x_{i,B})
    \prod_{r=1}^{l} p_{i,r}(\x_{i,r}, \D \x) \right)^{1/P^3} /Z,
\end{equation}
so that now $p_{i,r}$ represents a joint distribution between the local patch $\x_{i,r}$
\emph{and} the downsampled image $\D \x$, with $\D$ representing an operator that downsamples the whole image to a smaller image size.
Then 
\begin{align*}
    \log p(\x) &= \frac{1}{P^3} \Big( \sum_{i=1}^{P^3} (\log p_{i, B}(\x_{i,B})
    \\&
    + \sum_c \log p_c(\G_c \x, \D \x)) \Big),
\end{align*}
where for convenience we reindex the patches with $c$
and $\G_c$ denotes the same patch grabbing operator as aforementioned.
To take the gradient of this model,
first define $\bu = \G_c \x$ and $\bv = \D \x$.
Then 
\begin{align} \label{score_full}
    \nabla \log p(\x) &=
    \frac{1}{P^3} \left( \sum_{i=1}^{P^3} \sum_c (\G_c' \nabla_{\bu} \log p_c(\G_c \x, \D \x) \right.
    \nonumber \\ 
    &\quad + \left. \D' \nabla_{\bv} \log p_c(\G_c \x, \D \x)\right),
\end{align}
where we dropped the gradient of the bordering term since it is known to be zero,
and we applied the chain rule.
Hence, we can approximate the score functions
$\nabla_{\bu} \log p_c(\G_c \x, \D \x)$ and $\nabla_{\bv} \log p_c(\G_c \x, \D \x)$
with the neural networks $\s_u$ and $\s_v$ respectively.
The denoising score matching loss is then 
\begin{align} \label{joint_dsm}
    L(\boldsymbol{\theta}) = \frac{1}{P^3} & \mathbb{E}_{\bepsilon \sim \mathcal{N}(0, \sigma_t^2 \bI)} \|
    \sum_{i=1}^{P^3}  \sum_c
    (\G_c' \s_u (\bu, \bv;\boldsymbol{\theta})
    \nonumber\\
    &+ \D' \s_v (\bu, \bv;\boldsymbol{\theta})) - P^3 \bepsilon / \sigma_t^2 \|_2^2. 
\end{align}
Note that $\hat{\x} = \x + \bepsilon$ and now $\bu = \G_c \hat{\x}, \bv = \D \hat{\x}$,
and \x is sampled from the clean image distribution
but we dropped these terms from the expectation for simplicity. 

Unfortunately, due to the double summation inside of the norm, it is computationally infeasible to train the network using this loss function. Hence, we first use the inequality 
\begin{equation} \label{eq:blend_ineq}
    \| X_1 + \ldots + X_n \|_2^2 \le n \cdot (\| X_1 \|_2^2 + \ldots + \| X_n \|_2^2), 
\end{equation}
to upper bound the loss \eqref{joint_dsm} without the expectation to get 
\begin{align}
    L \le & C \sum_{i=1}^{P^3} \| \sum_c \G_c' \nabla_{\bu} \log p_c(\bu, \bv) \nonumber \\
    &+ \D' \nabla_{\bv} \log p_c(\bu, \bv) - \bepsilon / \sigma_t^2\|_2^2,
\end{align}
for a constant $C>0$.
Therefore, during training we can choose a random $i$ and then minimize the inner sum.
For fixed $i$, we usually choose each of the patches $c$ in a non-overlapping way. 
Unfortunately, the $\D'$ operator upscales the $\nabla_{\bv}$ patch-sized term into a whole image sized term,
so the sum over $c$ will contain contributions from each patch for every pixel.
However, the part that is outside the patch $c$ is difficult to learn
as the only contribution is from the upscaled part involving $\D'$. 
Thus we instead focus on learning the patch-specific part inside of the loss.
Rewriting the norm part of the loss, this is 
\begin{align} \label{eq: training_loss}
    L_c \approx & \| \sum_c \G_c \G_c' \nabla_{\bu} \log p_c(\bu, \bv) \nonumber \\
    &+ \G_c \D' \nabla_{\bv} \log p_c(\bu, \bv) - \G_c \bepsilon / \sigma_t^2\|_2^2.
\end{align}

When factoring out the $\G_c$ term,
we observe that each term in this sum independently contributes
to one of the non-overlapping patches of the entire image.
Hence, when training the network,
we do not have to perform denoising score matching across the entire image:
we need only add noise to the entire image,
provide one of the patches and the downsampled image to the network,
and have the network learn to denoise just the patch $\G_c \bepsilon$. 
In practice, to represent the joint score functions of $\G_c \x$ and $\D \x$, 
we input $\G_c \x$ and $\D \x$ to the network
via concatenation along the channel dimension 
(where $\D$ is chosen appropriately such that $\G_c \x$ and $\D \x$ are the same size),
and represent $\nabla_{\bu} \log p_c(\G_c \x, \D \x)+\nabla_{\bv} \log p_c(\G_c \x, \D \x)$
using single channels of the network output.

To facilitate the network to learn different prior distributions of 3D local patches at different locations, it is desirable to incorporate the positional information of the 3D patches.
We do this by defining the $x$ positional array
as a 3D array representing the $x$ coordinates
of each pixel in the image, and normalizing to be scaled between -1 and 1.
Similarly, $y$ and $z$ positional arrays are defined for $y$ and $z$ coordinates of the pixels, respectively.
To enable the network to learn patch distributions
that vary based on location within the image,
we extract patches from these positional arrays corresponding to the image patches.
As shown in Figure~\ref{fig:Training}, these positional patches are also concatenated along the channel dimension with $\G_c \x$ and $\D \x$. 

In summary, we use a 3D UNet with five input channels:
the noisy patch, the downsampled (noisy) image, and the three positional arrays.
The network has one output channel,
used to approximate $\nabla_{\bu} \log p_c(\bu, \bv)$ and $\nabla_{\bv} \log p_c(\bu, \bv)$,
and is trained according to the loss \eqref{eq: training_loss}. 
Most importantly, the spatial dimensions of the input are only in the 3D patch size of $P \times P \times P$ as shown in Figure~\ref{fig:Training}
Therefore, we never need to input the whole 3D image into the network. 
The appendix shows the pseudocode for the training algorithm.

\subsection{Sampling and Reconstruction Algorithm}

For the underlying sampling algorithm,
we choose DDIM \cite{DDIM} as a fast and reliable sampler
that many other works \cite{DDNM, DDRM, chung_ddip} have also used
as the backbone for solving inverse problems.
During the sampling step, each update uses \eqref{eq:DDS_DDIM_eq}.

The score function model \eqref{score_full}
allows for estimation of the score of the whole image from the trained network. 
However, it requires averaging over $P^3$ terms in the outer loop,
which would be prohibitively expensive. 
To address this issue,
\cite{hu2024learningimagepriorspatchbased} proposed a method that,
instead of taking the average,
chooses one of the terms at random and uses that as the approximation of the score.

However, when the number of sampling steps is limited,
this approach becomes inapplicable,
introducing boundary artifacts between patches.
Also, the sampling process on one initial point
does not consider the joint distribution of pixels across different patches at the same time step.
Here, we propose a novel recurrent noising strategy to further mitigate the boundary artifacts when reducing the sampling steps for improving sampling efficiency.
Specifically, at each iteration of DDIM, we denoise and renoise the image for $K$ times,
where in each time a different randomly chosen patch location, 
and finally average the results over $K$ times to obtain the sample.
The sampling equation is
\begin{equation}\label{eq: our_denoising}
    \x_{t-1} = \sqrt{\bar{\alpha}_{t-1} }\hat{\x}_t^{avg}+ 
    \sqrt{1-\bar{\alpha}_{t-1} -\sigma_t^2}\cdot\hat{\bepsilon}_t+\sigma_t\bepsilon_t
\end{equation}
where $\sigma_t$ is equal to \eqref{eq:DDS_sigma},
$\hat{\x}_t^{avg}$ is the averages of $\{\hat{\x}_{i^w,t}\}^K_{w=1}$
and $\hat{\bepsilon}_t$ is the scaled sum of $ \{\bepsilon_{i^w,t}\}^K_{w=1}$,
computed over $K$ randomly chosen patch locations $i^w$.
Here, noise variance is preserved similar to \eqref{eq:DDS_DDIM_eq},
allowing this method to sample from $q(\x_{t-1}|\x_0)$.

Previous works
\cite{chung2024decomposeddiffusionsampleraccelerating, song2024diffusionblendlearning3dimage}
showed that conjugate gradient method allows the model
to apply data consistency with measurement much faster.
At each timestep $t$,
using the denoised estimate \eqref{eq:denoisedxt}
we apply conjugate gradient method on the following forward operator \A:
\begin{equation}
    \hat{\x}_t'=\text{CG}(\A^*\A,\A^*\y,\hat{\x}_t,M)
\end{equation}
where the CG operator runs $M$ CG steps for the equation
$\A^* \y=\A^* \A \x$.
We apply this data consistency term after $K$ iterations
and then apply DDIM sampling to solve the inverse problem.
To summarize, our proposed sampling algorithm is shown in \ref{alg: recon}.
    

\begin{algorithm} 
\caption{3D Patch Diffusion Sampling with Recurrent Noising} 
\label{alg: recon}
    \begin{algorithmic}[1] 
    \Require Recurrent noising number $K$, $\A$, $\y$, $P$
    \State Initialize $\x_T \sim \mathcal{N}(0,\I)$
    \For{$t=N:1$}
        \For {$w = 1:K$} 
            \State Randomly select integer $i^w\in[1,P^3]$
            \State For $1\leq r\leq l$, extract 3D patches $\x_{i,r,t}$
            \For {$r=1:l$}
            \State $\hat{\x}_{i^w,r,t},  \bepsilon_{i^w,r,t}\leftarrow \D_\theta(\x_{i^w,r,t},\sigma_t)$ 
            \EndFor
            \State Concatenate all $\hat{\x}_{i^w,r,t}$ to entire volume $\hat{\x}_{i^w,t}$
            \State Concatenate all $\bepsilon_{i^w,r,t}$ to entire volume $\bepsilon_{i^w,t}$
            \State $\bepsilon \sim \mathcal{N}(0,\I)$
            \State $\x_t \leftarrow \sqrt{\bar{\alpha_t}}\hat{\x}_{i^w,t} +\sqrt{1-\bar{\alpha_t}}\bepsilon$ 
        \EndFor 
        \State $\hat{\x}_t^{\mathrm{avg}} = \sum\limits_{w=1}^{K} \hat{\x}_{i^w,t} / K$ , \; $\hat\bepsilon_t = \sum\limits_{w=1}^{K} \bepsilon_{i^w,t} / \sqrt{K}$ 
        \State Set $\hat{\x}_t'=\text{CG}(\A^*\A,\A^*\y,\hat{\x}_t^{\mathrm{avg}})$
        \State Sample $x_{t-1}$ via DDIM using \eqref{eq:DDIM_eq}
    \EndFor
    \end{algorithmic} 
    
\end{algorithm}
\section{Results}
\subsection{Experimental Setup}

\paragraph{Dataset preparation.}
We conducted experiments using the LIDC-IDRI dataset~\cite{lidc}.
The original dataset has a spatial resolution of $512 \times 512$ in the XY-plane,
with varying sizes along the Z-axis.
In our experiments, we selected the first 256 slices along the Z-axis
and filtered out samples with inconsistent shapes.
To ensure comparability with contemporary comparison methods,
we initially rescaled each XY-plane to $256 \times 256$.
This preprocessing resulted in 90 volumes for training and 1 volume for testing.
We also performed experiments on the AAPM dataset~\cite{AAPMct}
that contains only 10 CT volumes,
to evaluate our method’s performance with a limited amount of data.
This dataset was preprocessed in the same manner as the LIDC-IDRI dataset.
To demonstrate that our approach can be extended to larger volumes,
we additionally prepared a version of the LIDC-IDRI dataset
with clinically relevant dimensions of $ 512 \times 512 \times 256 $.

\paragraph{Baseline setting.}
For the LIDC-IDRI dataset,
the model was trained for approximately 8 days with a batch size of 64
and a patch size of $32 \times 32 \times 32$.
For the AAPM dataset, which contains only 9 volumes,
we trained the model from scratch for 5 days using the same batch and patch sizes.
For the larger LIDC-IDRI dataset,
the model was trained for 28 days with a batch size of 16
and a patch size of $64 \times 64\times32$.
All training was done on a single NVIDIA A100 GPU.
We applied our proposed sampling method using $K = 2$ and $\eta = 0.8$
for both $256 \times 256 \times 256$ and $ 512 \times 512\times256 $ volumes.
The total number of model parameters is 68.59 million.

\subsection{Unconditional 3D Image Generation}
\begin{figure*}[ht]
    \centering
    \includegraphics[width=1.0\linewidth]{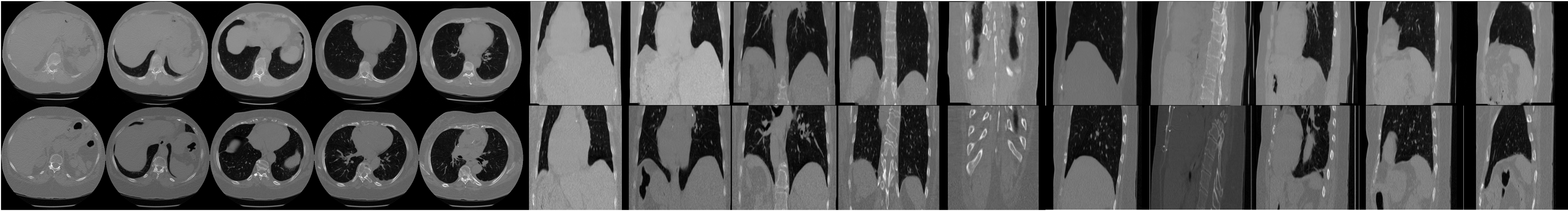}
    \caption{Unconditional 3D image generation results using the LIDC-IDRI prior. The top row shows axial, coronal, and sagittal slices from a generated volume, and the bottom row shows the corresponding slices from its nearest-neighbor volume in the training dataset. The slice indices for the axial, coronal, and sagittal views are $[30, 80, 130, 180, 230]$, $[70, 100, 130, 160, 190]$, and $[60, 130, 160, 190, 210]$, respectively}
    \label{fig:generation result 3d}
\end{figure*}

Our proposed method is able to learn a fully 3D prior
on $256\times256\times256$ sized CT volumes,
allowing it to generate high quality volumes. 
We used \eqref{eq:DDS_DDIM_eq} with $K = 1$ and $\eta=0.4$ and 200 steps for unconditional sampling.
Figure~\ref{fig:generation result 3d} shows unconditional volume samples from LIDC-IDRI prior,
visualized across axial, coronal, and sagittal slices. 
The top row shows slices from an unconditionally sampled volume, while the bottom row shows the nearest-neighbor volume from the training dataset.
These results indicate that the proposed prior does not simply memorize from the training dataset and is capable of generating realistic high-resolution 3D CT volumes with fine-grained details of anatomic structure.
Further generation results on different prior are in the appendix.

\begin{table*}[ht!]
\centering
\caption{PSNR (dB) comparison of various methods for CT reconstruction on different datasets and resolutions with best results in bold.}
\label{psnr_all}
\begin{tabular}{c|ccc|ccc|ccc}

\toprule
Dataset / Method & \multicolumn{3}{c|}{LIDC ($256\times256\times256$)} & \multicolumn{3}{c|}{LIDC ($512\times512\times256$)} & \multicolumn{3}{c}{AAPM ($256\times256\times256$)} \\
\midrule
Views & 8 & 20 & 60 & 8 & 20 & 60 & 8 & 20 & 60 \\
\midrule
FBP & 15.08 & 20.21 & 29.37 & 14.46 & 20.05 & 28.02 & 13.81 & 18.39 & 27.71 \\
ADMM-TV & 21.15 & 24.12 & 29.43 & 22.47 & 25.39 & 29.38 & 23.71 & 26.75 & 31.28 \\
FBP-UNet \cite{fbpconvnet} & 22.60 & 27.77 & 32.78 & 26.09 & 30.32 & 35.34 & 25.93 & 30.06 & 36.93 \\
DDS \cite{chung2024decomposeddiffusionsampleraccelerating} & 23.23 & 29.92 & 35.79 & 24.37 & 27.33 & 28.61 & 30.18 & 34.94 & 38.87 \\
DiffusionBlend \cite{song2024diffusionblendlearning3dimage} & 30.43 & 35.89 & 40.87 & 31.69 & 35.94 & 39.32 & -- & -- & -- \\
Blend + FT \cite{song2024diffusionblendlearning3dimage} & 32.94 & 37.05 & 42.72 & 33.01 & 36.44 & 39.31 & 32.25 & 37.21 & 42.21 \\
\textbf{Proposed} & \textbf{33.06} & \textbf{38.56} & \textbf{43.70} & \textbf{33.17} & \textbf{37.33} & \textbf{40.16} & \textbf{32.96} & \textbf{38.47} & \textbf{42.66} \\
\bottomrule
\end{tabular}
\end{table*}
\subsection{Solving Inverse Problems}
\begin{figure*}[ht!]
\centering
\includegraphics[width=0.9\linewidth]{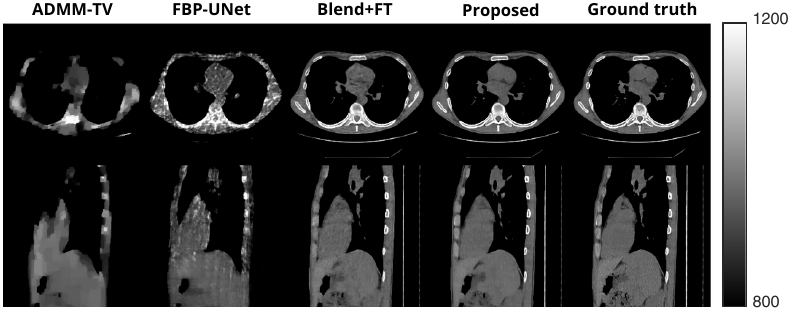}
\caption{Results of our proposed method and comparison methods
for 20 view CT recon on LIDC $256 \times 256\times 256$ dataset.
Images are shown in modified Hounsfield units.
The top row shows the axial slice
and the bottom row shows the sagittal slice from the reconstructed volume.}
\label{fig:lidc20}
\end{figure*}

We conduct experiments of sparse-view CT (SVCT) reconstruction
from 8, 20, and 60 views on one of the test volumes. 
The results in Table~\ref{psnr_all} and Figure~\ref{fig:lidc20} demonstrate that our method is able to reconstruct high-quality images.
Furthermore, we compared our proposed method with various classical methods
and diffusion model methods for 3D CT reconstruction.
We used the filtered back projection method with a ramp filter,
whose implementation is found in \cite{chung:22:s3i}.
We also used the total variation (TV) regularizer 
and solved the optimization problem using ADMM (ADMM-TV)
with the implementation in \cite{Hong_2024}.
For the other baseline, we implemented FBP-UNet \cite{fbpconvnet}
which is a supervised method that involves training a UNet
that maps FBP reconstruction to the clean image.
Since FBP-UNet is a 2D method,
we learned a mapping between 2D slices and then stacked the 2D slices
to get the final 3D volume.

For diffusion model methods,
we ran DDS \cite{chung2024decomposeddiffusionsampleraccelerating}
and DiffusionBlend \cite{song2024diffusionblendlearning3dimage}. 
To obtain the best possible results for DiffusionBlend, 
we used the trick in the original work
that involved taking a network that was pretrained on ImageNet
and then finetuning it on the relevant CT dataset. 
Thus, we specifically ran DiffusionBlend++ with the prior trained on groups of 3 slices. 
To illustrate the effect of using the pretrained network,
we show experiments both with the finetuning method (Blend+FT)
and with networks that were trained from scratch on the CT datasets.
Since our proposed method does not rely on a pretrained network,
the most fair comparison is with \cite{song2024diffusionblendlearning3dimage}
without a pretrained network;
nevertheless we showed both results
to illustrate how our proposed method can obtain superior results.
Both of these methods were run with 200 sampling steps.
The appendix provides the experiment parameters.

\begin{table}[ht!]
\centering
\begin{center}
\caption{Per volume runtimes of different methods
for 8 view CT recon.}
\label{runtimes}
\begin{tabular}{cc}
\toprule
Method & Runtime (minutes) $\downarrow$ \\
\midrule
FBP  & 0.1 \\
ADMM-TV  & 6 \\
FBP-UNet  & 1 \\
DDS & 64 \\
DiffusionBlend & 55 \\ 
Proposed & 20 \\
\bottomrule
\end{tabular}
\end{center}
\end{table}

\subsection{Ablation studies}
We conducted ablation studies to analyze the factors
that may influence the performance of our 3D patch diffusion model.

\paragraph{Impact of downsample channel.}
Downsample channel plays a critical role in our method,
allowing the model to learn the global structure.
Without the downsample channel the model will only learn the local relationship between patches,
failing to generate realistic CT images.
To validate this, we compared the generation quality of the 3D patch-based diffusion model with and without the downsample channel using the Fréchet Inception Distance (FID), as shown in Fig.~\ref{fid_lidc256}.

\begin{table}[ht!] 
\centering 
\begin{center} 
\caption{FID comparison with and without downsample channel} 
\label{fid_lidc256} 
\begin{tabular}{c|cc} 
\toprule Method & w. downsample & w.o. downsample  \\ 
\midrule FID $\downarrow$ & 40.80 & 112.12 \\ 
\bottomrule
\end{tabular} 
\end{center} 
\end{table}


\paragraph{Impact of patch size.}
Consider an extreme case where the patch size is $1 \times 2$.
According to our modeled prior distribution in~\eqref{eq:padis_distribution},
using such small patches is theoretically valid but
unlikely to fully capture image statistics.
To better understand this behavior,
we examined how patch size affects the model’s performance on inverse problem tasks.
We trained models with varying patch sizes,
which is feasible due to the input-size-agnostic property of U-Net architectures.
Patch sizes were chosen by dividing the full volume size by a scaling factor.
For example, a scaling factor of 8 on a $256 \times 256 \times 256$ volume
corresponds to patches of size $32 \times 32 \times 32$.
Since the data are volumetric,
doubling the patch size in each dimension reduced the batch size by a factor of 8.
As a result, smaller patches train faster in wall-clock time:
the factor-16 setting converges within 3 days,
whereas the factor-8 setting requires approximately 8 days on $256 \times 256 \times 256$ volumes.

Experimental results indicate that as the patch size decreases,
the model failed to capture meaningful semantic structure,
leading to reduced PSNR on the sparse-view CT (SVCT) task.
As shown in Table~\ref{patchsize}, model performance degrades as the scaling factor increases.
This observation suggests that when patches become too small relative to the full image,
they behave more like individual pixels.
Consequently, the model can no longer learn a coherent image-level prior,
causing~\eqref{eq:padis_distribution} to describe a less effective model.

\begin{table}[ht!]
\centering
\begin{center}
\caption{PSNR in dB of different patch and volume size}
\label{patchsize}
 \begin{tabular}{c|cc}
 \toprule
 Factor & $256\times256\times256$ & $ 512 \times 512\times256 $\\
 \midrule
  8  & \textbf{33.06} & \textbf{33.17}\\
 16  & 27.87 & 31.71\\
 \bottomrule
 \end{tabular}
 \end{center}
\end{table}

\paragraph{Impact of recurrent noising during sampling process.}
This section explores how recurrent noising affects the reconstruction quality.
To understand how the number of iterations $K$ affects the reconstructed images,
we tested on the 8-view SVCT task
with 200 sampling steps.
Table~\ref{K_test} shows that recurrent noising during sampling
($K > 1)$
yielded higher PSNR than reconstruction without any recurrent noising ($K=1$).
The PSNR peaked at $K=2$
and gradually decreased as $K$ increased.
We conjecture that this is because the estimated $\hat\bepsilon_t$
is not strictly sampled from a Gaussian distribution,
causing error accumulation as $K$ increases.
\begin{table}[ht!]
\centering
\begin{center}
\caption{PSNR in dB of different $K$, with best results in bold}
\label{K_test}
 \begin{tabular}{c|c|c|c}
 \toprule
 $K$ & 8 view & 20 view & 60 view \\
 \midrule
 1 & 32.53 & 38.18 & 43.46\\
 2 & \textbf{33.06} & \textbf{38.56} & \textbf{43.70} \\
 3 & 32.97 &38.43 & 43.52 \\
 4 & 32.74 &38.15& 43.44 \\ 
\bottomrule
\end{tabular}
\end{center}
\end{table}



\section{Conclusion}
In this work,
we propose a novel 3D patch-based diffusion model
that is conditioned with downsampled volume and coordinates to enable learning the 3D CT image prior in an efficient way.
By learning from the local patches coupled with global context, our model is able to generate high-resolution 3D images.
Experiments across different volume sizes and CT datasets
showed that our method achieve SOTA results on SVCT reconstruction
while reducing inference time by more than $2\times$.
One of the limitations
is that our approach assumes the data is well aligned to a consistent coordinate,
which may reduce applicability to less structured image domains.
In the future, we plan to investigate alternative methods that allow the model to learn the positional information of the patch, thereby improving robustness in less structured data.

\section*{Acknowledgement}
LS acknowledges funding support by
NSF (National Science Foundation) via grants IIS-2435746,
Defense Advanced Research Projects Agency (DARPA) under Contract No. HR00112520042,
Hyundai America Technical Center, Inc. (HATCI),
as well as the University of Michigan MICDE Catalyst Grant Award and MIDAS PODS Grant Award.
JH acknowledges the J. Robert Beyster Computational Innovation Graduate Fellowship.

{
\small
\bibliographystyle{unsrtnat}
\bibliography{main}
}

\appendix 

\clearpage
\setcounter{page}{1}
\maketitlesupplementary


\section{Training Algorithm}
Algorithm \ref{alg: train}
summarizes
the training algorithm for 3D patch diffusion.

\begin{algorithm} 
\caption{Training 3D patch diffusion model} \label{alg: train}
    \begin{algorithmic}[1] 
    \Require image size $N_x\times N_y\times N_z$, patch size $P$
    \Repeat 
    \State $t\sim \text{Uniform}(\{ 1, ...T\})$
    \State $c\sim \text{Uniform}(\{0,...(N_x+P)(N_y+P)(N_z+P)\})$
    \State $\bepsilon\sim \mathcal{N}(0,\I)$
    \State $\x_t = \sqrt{\bar\alpha_t}\x+\sqrt{1-\bar\alpha_t}\bepsilon$
    \State $\u = \G_c \x_t$ \Comment{Patchify}
    \State $\v = \D \x_t$ \Comment{Downsample}
    \State Apply gradient descent on $\nabla_\theta||D_\theta(\u,\v)-\G_c\bepsilon||^2$
    \Until {converged}
    \end{algorithmic} 
\end{algorithm}

\section{Experiment Parameters}
For the network used in the proposed method,
we extended the 2D UNet architecture used in~\cite{DDIM} to 3D
by replacing all 2D operators with their 3D counterparts. 
The model was optimized by Adam,
and we used an exponential moving average with a decay rate of $0.999$
to stabilize the training.
The diffusion process used $1000$ timesteps
with a linearly increasing noise variance schedule from $0.0001$ to $0.02$.
All of our proposed models were trained with the same architecture.
Table~\ref{hyperparams}
summarizes the architecture and hyperparameter settings.

\begin{table}[ht]
\centering
\caption{Model hyperparameters.}
\label{hyperparams}
\begin{tabular}{cc}

\toprule
Base channel width & 64\\
Channel multipliers & $[1,2,4,4]$  \\
\# of input channel & 5  \\
\# of output channel & 1  \\
Attention resolution & $[8, 8]$\\
\# of residual block & 2 \\
Learning rate & $2 \times 10^{-5}$ \\

\bottomrule

\end{tabular}
\end{table}

\section{Limitations}

\begin{figure}[ht]
    \centering
    \includegraphics[width=0.8\linewidth]{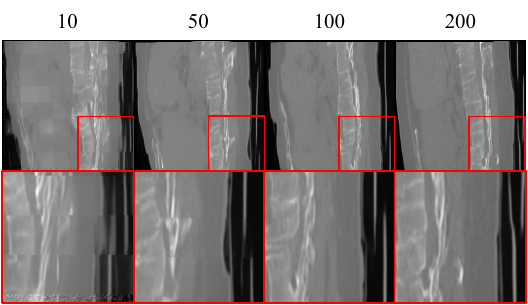}
    \caption{Sagittal slices of generated 3D volumes
    for different numbers of different sampling steps.
    Generated image quality degrades if the number of steps
    is reduced too much.
    }
    \label{fig:patch_artifact}
\end{figure}

\paragraph{Limitation on reducing sampling steps}
In Eq.~\ref{joint_down}, we model our 3D prior as the product of all non-overlapping patches for a given initial point, and then take the product over all possible initial points.
However, during each sampling step, we approximate the 3D prior as the product of all non-overlapping patches from one random initial point~\cite{hu2024learningimagepriorspatchbased} to reduce the computation cost.
Consequently, the model cannot be sampled with smaller step sizes, as it relies on a random initial point at each step to reduce the boundary artifacts. 
Figure~\ref{fig:patch_artifact} shows that with smaller sampling steps, the boundary artifacts of patches still remains. 

\begin{figure}[ht]
    \centering
    \includegraphics[width=0.8\linewidth]{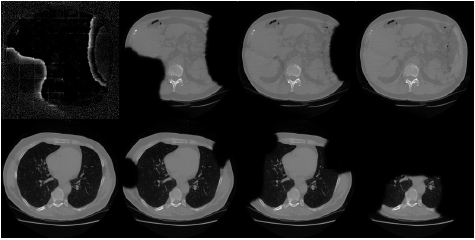}
    \caption{Top row shows the first four axial slices, and the bottom row shows the last four axial slices of the generated volume}
    \label{fig:boundary_limitation}
\end{figure}

\paragraph{Limitation on boundary region of volume}
Our 3D prior model
zero pads
each 3D volume.
The zero padding can create
unwanted artifacts
in the top few and bottom few slices
of generated volumes.
Figure~\ref{fig:boundary_limitation} shows the first and last four axial slices of a representative generated volume.
Future work could consider other padding approaches.

\paragraph{Low generation quality on large volume} 
Although our proposed method achieves SOTA performance on reconstructing $512\times512\times256$ CT volume, it fails to generate fine-grained details as shown in Figures~\ref{fig:gen_axial},~\ref{fig:gen_coronal},~\ref{fig:gen_sagittal}. We speculate that because the number of model parameters remains the same when training priors on $256\times256\times256$ and $512\times512\times256$ sized volumes, it is difficult to achieve comparable generative performance for the larger volume size.

\section{Trade-offs}
In this section, we discuss the trade-offs between the different patch sizes. All experiments were conducted on the $256\times256\times256$ LIDC-IDRI prior for the 8-view SVCT task. Here, the patch size plays a critical role in training efficiency: when the patch size doubles, then the batch size decreases by a factor of eight. However, if we reduce the patch sizes too much, the performance drops drastically, as shown in Figure~\ref{recon_diff_patch}. Therefore, we choose a patch size that is a factor of eight of the full image resolution as our baseline, balancing training efficiency and reconstruction quality. 
\begin{table}[ht]
\centering
\caption{Trade-offs on different patch sizes}
\label{recon_diff_patch}
\begin{tabular}{c|ccc}

\toprule
Patch size (factor) & PSNR $\uparrow$ & training time & sampling time \\
\midrule
$64\times64\times64$ (4) & 31.74 & 8 days & 30 min\\
$32\times32\times32$ (8) & \textbf{33.06} & 8 days & 20 min \\
$16\times16\times16$ (16) & 27.87 & 3 days & 10 min \\

\bottomrule

\end{tabular}
\end{table}

\onecolumn
\clearpage

\section{Additional Figures}
\begin{figure*}[ht]
    \centering
    \includegraphics[width=1.0\linewidth]{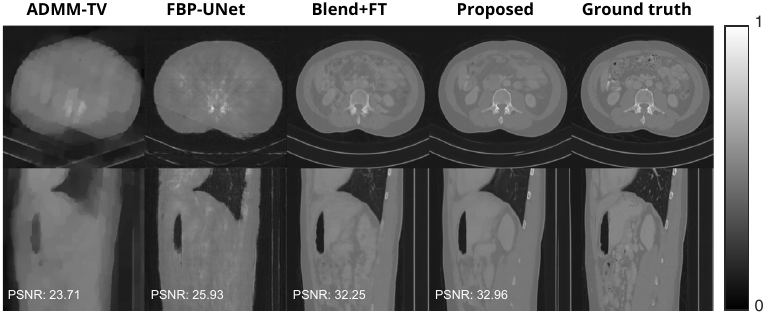}
    \caption{Results of our proposed method and comparison methods
    for 8-view 3D CT reconstruction on AAPM dataset.
    The top row shows an axial slice and the bottom row shows a sagittal slice
    from each reconstructed volume.}
    \label{fig:main_aapm8}
\end{figure*}

Figure \ref{fig:main_aapm8} shows visual results of 8-view CT reconstruction
on the AAPM dataset with our proposed method and some comparison methods.
Compared to the result with DiffusionBlend (fine-tuned from a checkpoint pretrained on natural images),
our method is able to preserve more details,
such as bone artifacts and blood vessels of a lung.

\begin{figure*}[ht]
    \centering
    \includegraphics[width=1.0\linewidth]{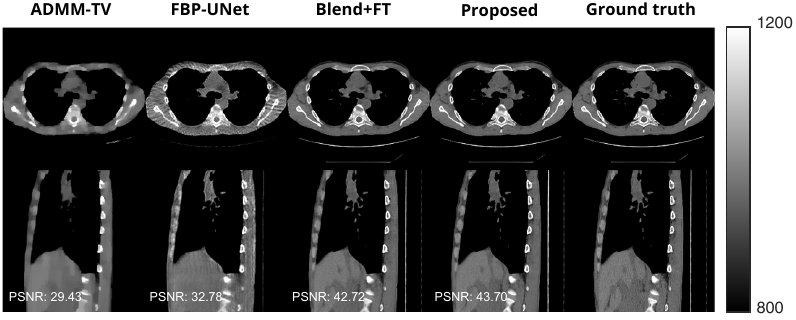}
    \caption{Results of our proposed method and comparison methods
    for 60 view CT recon on $256 \times 256\times 256$ LIDC dataset.
    Plots are shown in Hounsfield units.
    The top row shows the axial slice and the bottom row shows the sagittal slice from the reconstructed volume.}
    \label{fig:lidc60}
\end{figure*}

\noindent
Figure \ref{fig:lidc60} shows visual results of 60-view CT reconstruction
on the $256 \times 256\times 256$ LIDC dataset with our proposed method and some comparison methods.
We see that even in this fairly sparse setting, the proposed method is able to obtain a very high quality reconstruction that is absent of any hallucinations or artifacts.

\begin{figure*}[ht]
    \centering
    \includegraphics[width=1.0\linewidth]{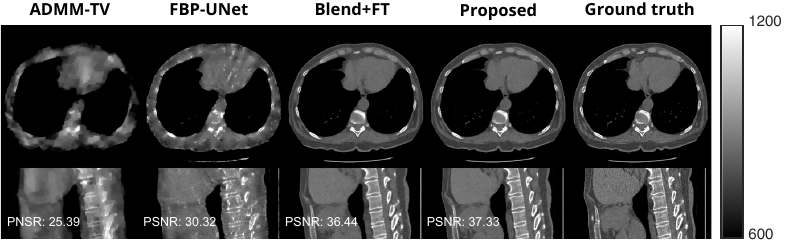}
    \caption{Results of our proposed method and comparison methods
    for 20 view CT recon on $512 \times 512\times 256$ LIDC dataset.
    Plots are shown in Hounsfield units.
    The top row shows the axial slice and the bottom row shows the sagittal slice from the reconstructed volume.}
    \label{fig:lidc512_20}
\end{figure*}

\begin{figure*}[ht]
    \centering
    \includegraphics[width=1.0\linewidth]{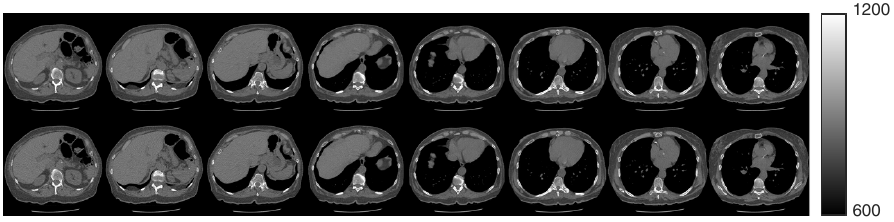}
    \caption{Results of our proposed method for 60 view CT recon on $512 \times 512\times 256$ LIDC dataset.
    Plots are shown in Hounsfield units.
    The top row shows our proposed method and the bottom row shows the ground truth. All slices are axial slices from the same volume.}
    \label{fig:lidc512_many}
\end{figure*}

\clearpage
\noindent
Figure~\ref{fig:lidc512_20},~\ref{fig:lidc512_many} shows visual results of 20-view and 60-view CT reconstruction
on the $512 \times 512 \times 256$ LIDC dataset with our proposed method and some comparison methods. 
To the best of our knowledge, we are the first work to leverage unconditional diffusion priors to perform CT reconstruction on CT volumes of size 512. 

\begin{figure*}[ht]
    \centering
    \includegraphics[width=1.0\linewidth]{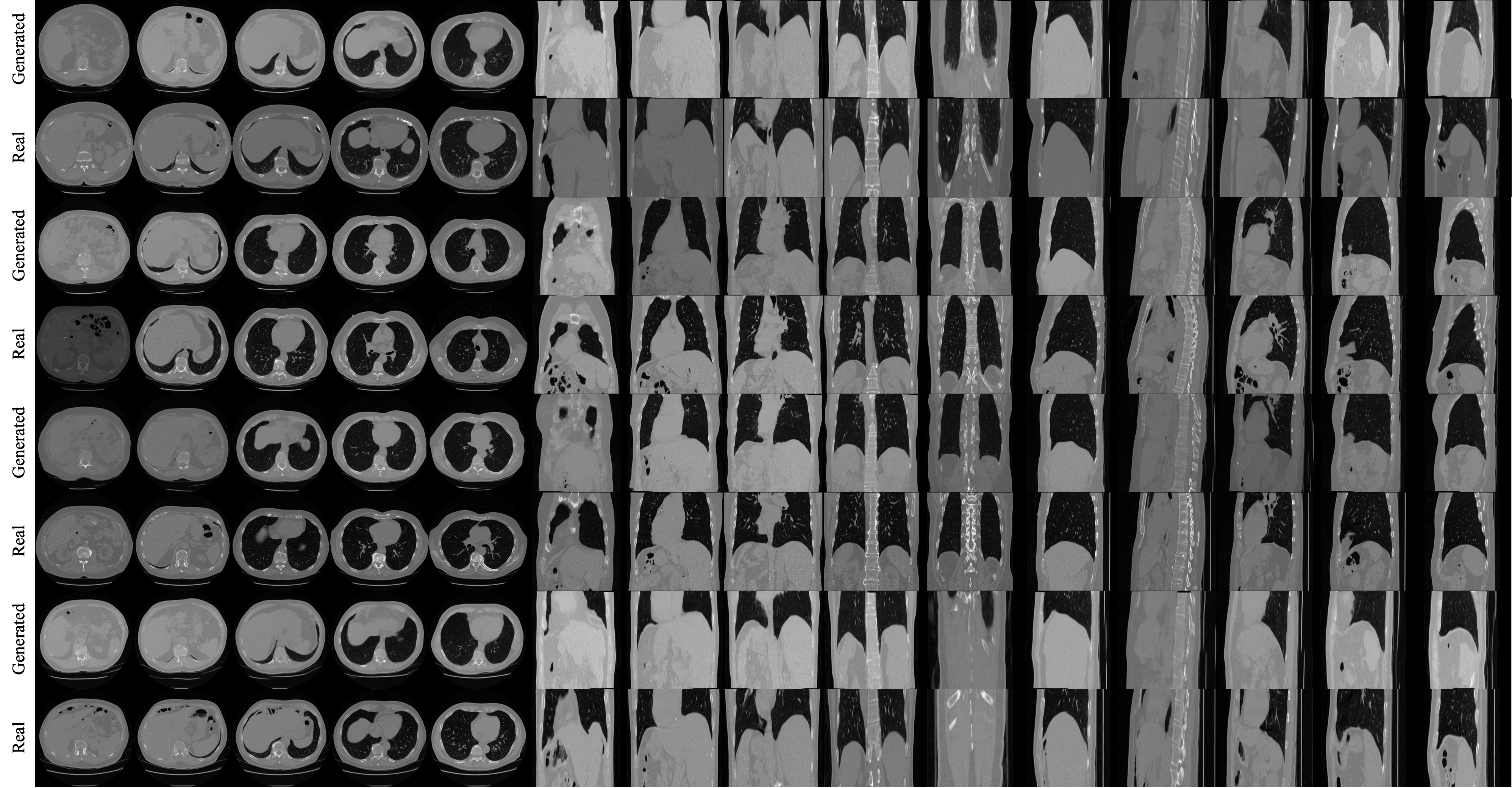}
    \caption{Unconditional sampling result on $256\times 256\times 256$ LIDC-IDRI prior and its nearest neighbor from the training dataset sliced by axial, coronal, and sagittal view.}
    \label{fig:lidc256_gen_nn}
\end{figure*}

\begin{figure*}[ht]
    \centering
    \includegraphics[width=1.0\linewidth]{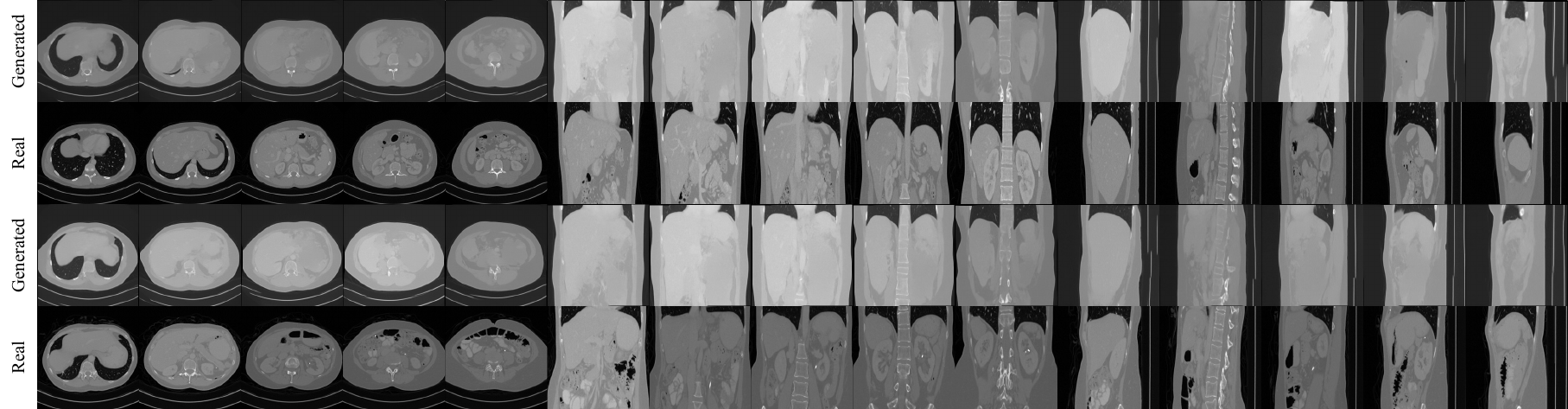}
    \caption{Unconditional sampling result on $256\times 256\times 256$ AAPM prior and its nearest neighbor from the training dataset sliced by axial, coronal, and sagittal view.}
    \label{fig:aapm256_gen_nn}
\end{figure*}

\clearpage
\noindent
Figure~\ref{fig:lidc256_gen_nn}, ~\ref{fig:aapm256_gen_nn} shows the selected slice of unconditionally sampled volume from $256\times256\times256$ LIDC-IDRI prior and $256\times256\times256$ AAPM prior. We calculated its nearest neighbor from the training dataset and showed the corresponding slices below. This further solidifies that our model could generate volumes instead of memorizing from the training dataset. 

\begin{figure*}[ht]
    \centering
    \includegraphics[width=0.9\linewidth]{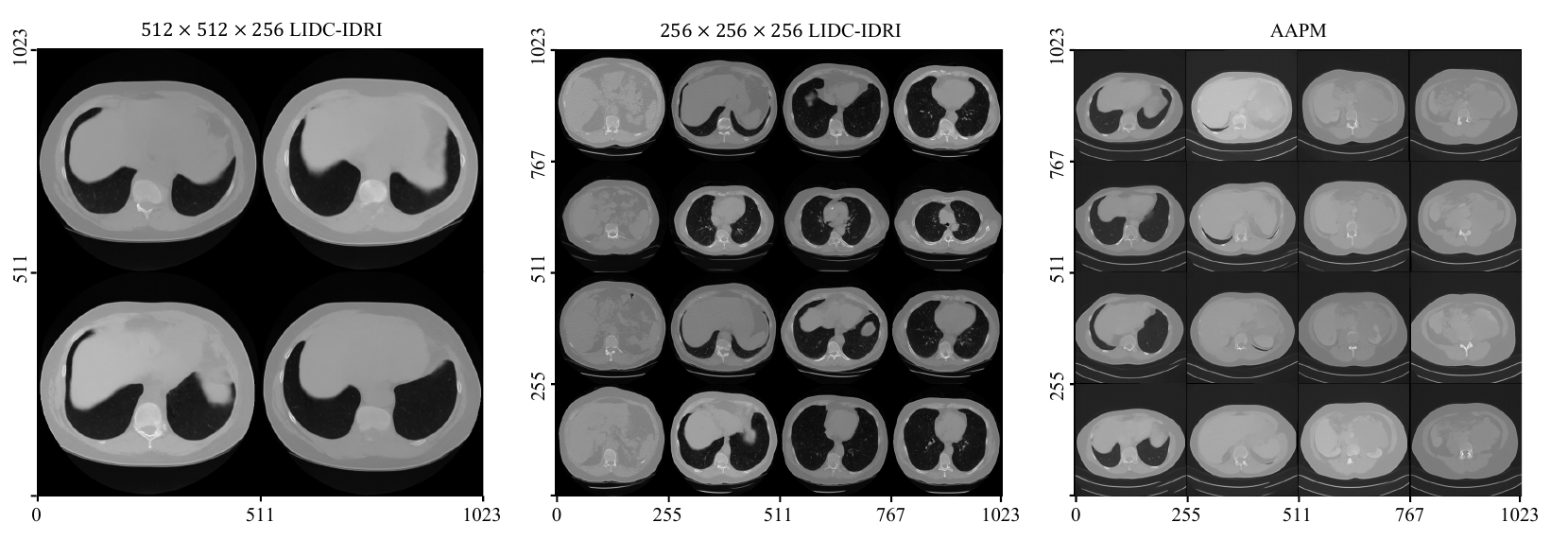}
    \caption{Unconditional sampling examples for three different priors
    (axial slices of 3D volumes).}
    \label{fig:gen_axial}
\end{figure*}

\begin{figure*}[ht]
    \centering
    \includegraphics[width=0.9\linewidth]{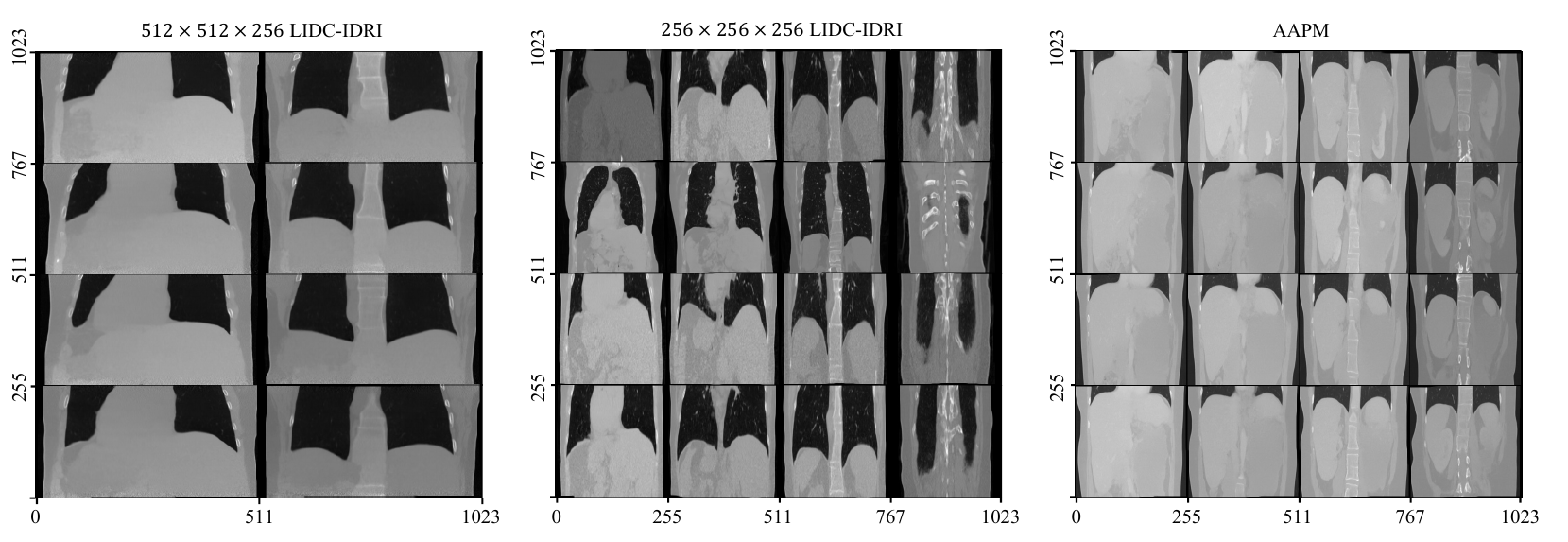}
    \caption{Unconditional sampling examples for three different priors
    (coronal slices of 3D volumes).}
    \label{fig:gen_coronal}
\end{figure*}

\begin{figure*}[ht]
    \centering
    \includegraphics[width=0.9\linewidth]{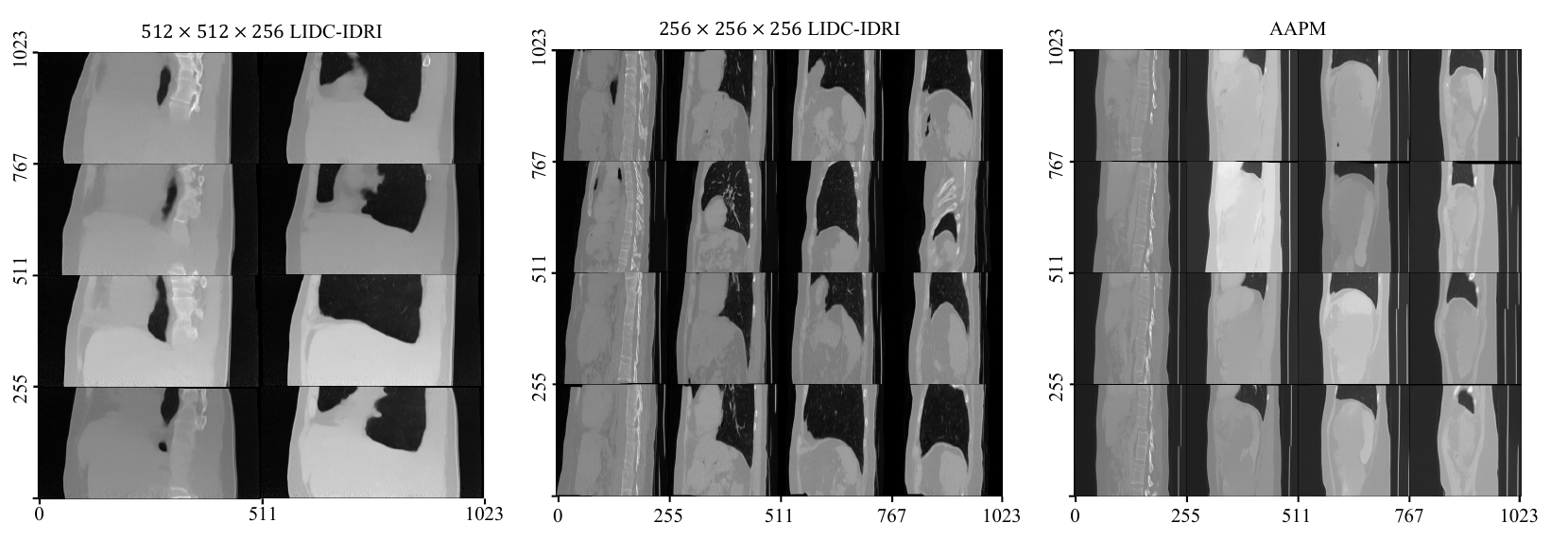}
    \caption{Unconditional sampling examples for three different priors
    (sagittal slices of 3D volumes).}
    \label{fig:gen_sagittal}
\end{figure*}

\clearpage
\noindent
Figures~\ref{fig:gen_axial},~\ref{fig:gen_coronal},~\ref{fig:gen_sagittal}
show unconditional sampling results of $512\times512\times256$, $256\times256\times256$ LIDC-IDRI prior and $256\times256\times256$ AAPM prior. 
Here, our proposed method is capable of generating realistic volumes,
even from the $256\times256\times256$ AAPM dataset that consists of only 9 training volumes. 

\begin{figure*}[ht]
    \centering
    \includegraphics[width=1.0\linewidth]{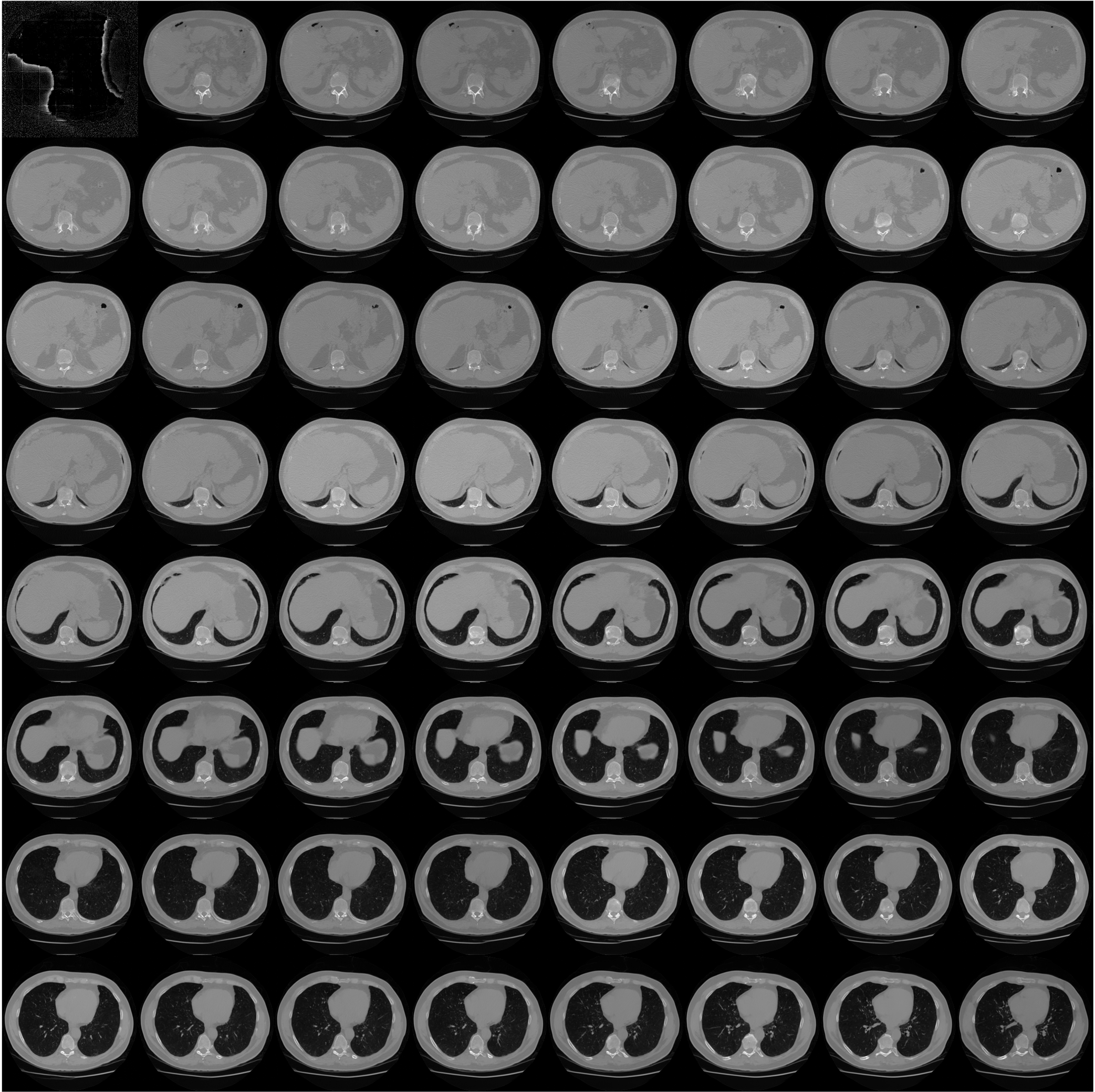}
    \caption{Continuous axial slice of a volume sampled from $256\times 256\times 256$ LIDC-IDRI prior; every fourth slice is shown}
    \label{fig:lidc_gen_256slice}

\end{figure*}

\begin{figure*}[ht]
    \centering
    \includegraphics[width=1.0\linewidth]{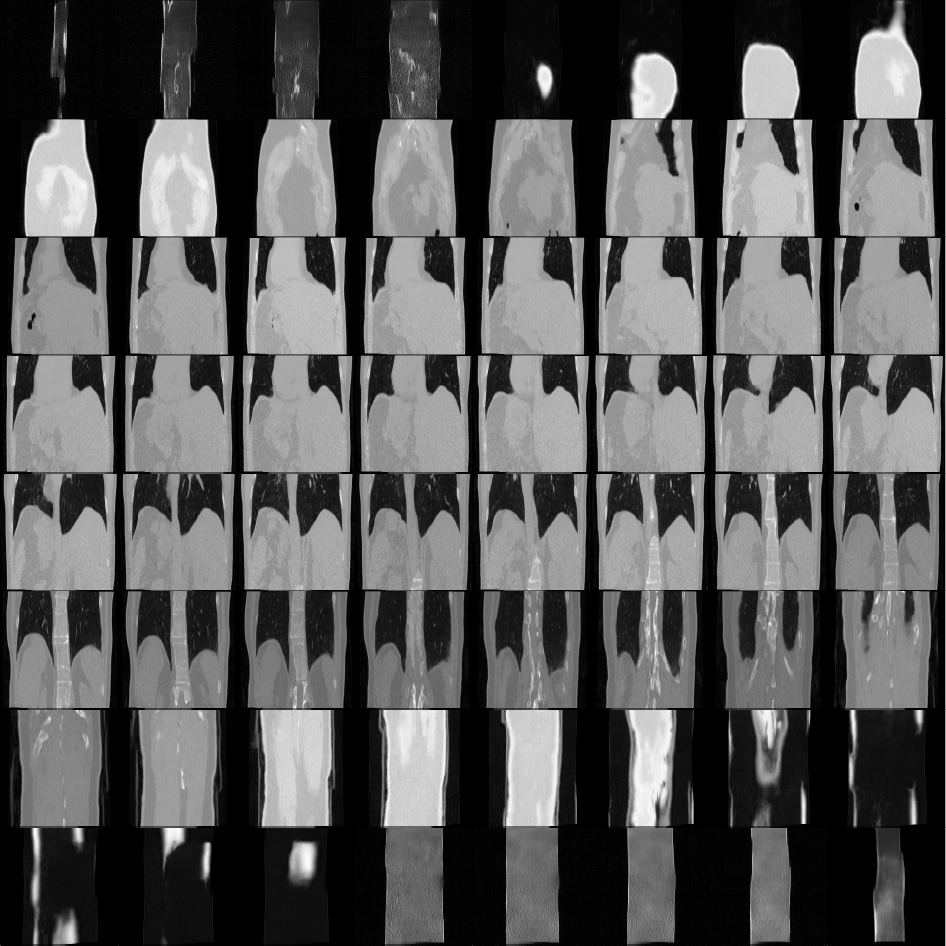}
    \caption{Continuous coronal slice of a volume sampled from $256\times 256\times 256$ LIDC-IDRI prior; every fourth slice is shown}
    \label{fig:lidc_gen_256slice_coronal}

\end{figure*}

\begin{figure*}[ht]
    \centering
    \includegraphics[width=1.0\linewidth]{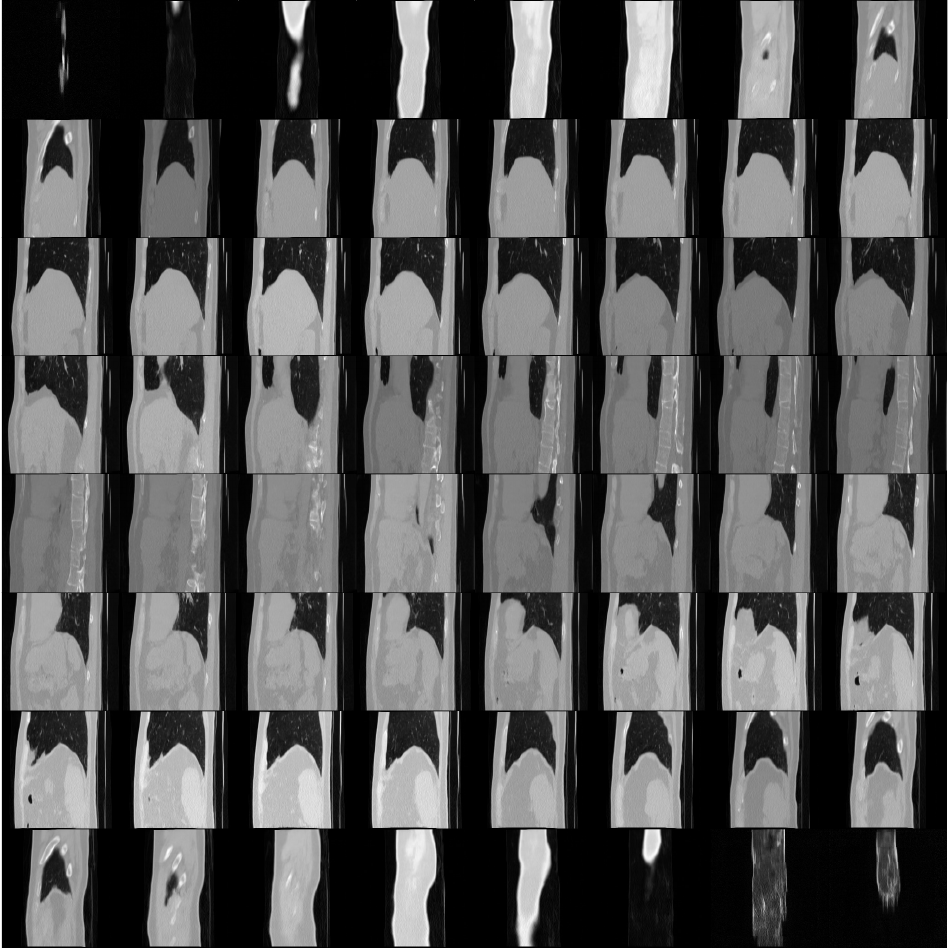}
    \caption{Continuous sagittal slice of a volume sampled from $256\times 256\times 256$ LIDC-IDRI prior; every fourth slice is shown}
    \label{fig:lidc_gen_256slice_sagittal}

\end{figure*}

\clearpage
\noindent
Figure~\ref{fig:lidc_gen_256slice},~\ref{fig:lidc_gen_256slice_coronal},~\ref{fig:lidc_gen_256slice_sagittal} shows the continuous axial, coronal and sagittal slices of a volume that is unconditionally sampled from $256\times256\times256$ LIDC-IDRI prior.

\end{document}